**Title**:

Survival Prediction from Imbalance colorectal cancer dataset using hybrid sampling methods and tree-based classifiers

**Authors' Names and affiliations**:

Sadegh Soleimani[1], Mahsa Bahrami[1], and Mansour Vali[1]

[1]Department of Biomedical Engineering, K. N. Toosi University of Technology, Tehran, Iran.

**Corresponding author**:

Mansour Vali

Department of Biomedical Engineering, K. N. Toosi University of Technology, Tehran, Iran, email: mansour.vali@eetd.kntu.ac.ir, Postal Code: 1631714191, P. O. Box: 16315-1355, Phone number: +98(21)8846-2174

**Abstract**

Background and Objective: Colorectal cancer is a high mortality cancer, with a mortality rate of 64.5% for all stages combined. Clinical data analysis plays a crucial role in predicting the survival of colorectal cancer patients, enabling clinicians to make informed treatment decisions. However, utilizing clinical data can be challenging, especially when dealing with imbalanced outcomes, an aspect often overlooked in this context. This paper focuses on developing algorithms to predict 1-, 3-, and 5-year survival of colorectal cancer patients using clinical datasets, with particular emphasis on the highly imbalanced 1-year survival prediction task. To address this issue, we propose a method that creates a pipeline of some of standard balancing techniques to increase the true positive rate. Evaluation is conducted on a colorectal cancer dataset from the SEER database, which exhibits high imbalance in the 1-year survival analysis and an imbalance in the 3-year analysis, achieving balance in the 5-year analysis. Methods: The pre-processing step consists of removing records with missing values and merging categories with less than 2% share for each categorical feature to limit the number of classes of each component. The minority class of 1-year and 3-year survival tasks consists of 10% and 20% of the data, respectively. Edited Nearest Neighbor, Repeated edited nearest neighbor (RENN), Synthetic Minority Over-sampling Techniques (SMOTE), and pipelines of SMOTE and RENN approaches were used and compared for balancing the data with tree-based classifiers. Five well-known classifiers were used for classification: Decision Trees, Random Forest, Extra Tree, eXtreme Gradient Boosting, and Light Gradient Boosting (LGBM) Method. Results: The performance evaluation utilizes a 5-fold cross-validation approach. In the case of highly imbalanced datasets (1-year), our proposed method with LGBM outperforms other sampling methods with the sensitivity of 72.30%. For the task of imbalance (3-year survival), the combination of RENN and LGBM achieves a sensitivity of 80.81%, indicating that our proposed method works best for highly imbalanced datasets. Additionally, when predicting 5-year survival, the sensitivity reaches 63.03% using LGBM. Conclusions: Our proposed method significantly improves mortality prediction for the minority class of colorectal cancer patients. RENN followed by SMOTE yields better sensitivity in the classifiers, with LGBM as the predictor performing best for 1- and 3-year survival. In the 5-year task, LGBM outperforms other models in terms of f-score.

**Keywords:** Colorectal cancer, repeated edited nearest neighbor, synthetic minority over-sampling, survival prediction, SEER.

# 1. Introduction

Colorectal cancer (CRC) is common cancer [1]. In fact, it had the fourth highest diagnosis and the third highest cancer-based death rank worldwide in 2018 [2]. Risk factors for developing CRC include smoking, obesity, unhealthy lifestyle, and alcohol consumption [3]. A confident prediction of each case's survivability would be helpful to health maintenance and can help physicians have a fair approximation of patients' survivability [4]. This information helps in making better decisions for patients' treatment.

Clinical data analysis provides a unique opportunity to study the importance of collected features and analyze patients' data. Machine learning algorithms can be used to achieve this goal [5]. Despite the high prevalence and death rates of CRC worldwide, it seems to be overlooked in such analysis and research compared to breast cancer or lung cancer [6].

Clinical data processing consists of several stages that involve collecting and cleaning patient data, storing it in a secure and organized manner, analyzing the data using statistical and machine learning methods, interpreting the results, and sharing and disseminating them to relevant parties [7]. The pre-processing stage involves imputing incomplete data, removing outliers, and normalizing data for further analysis. Feature selection aim to improve the performance of machine learning algorithms by selecting most discriminative features for the task, thus decreasing computational complexity and making algorithms more cost-efficient. Sampling and balancing techniques are mainly used to improve the capability of models to have better results on classification.

Machine learning or data mining algorithms classify data samples into different categories by extracting discriminative patterns from the data. These algorithms use pre-processed data that has gone through feature selection or dimension reduction to improve classification accuracy. By using various statistical and machine learning techniques, clinical data processing can extract valuable insights from patient data that can be used to inform clinical practice and health policy.

Although each step of data analysis has numerous methods to implement and utilize, knowing each method's capabilities and shortcomings can help to choose procedures most related to the work. Exhaustively searching for the best ways in each section would be time-consuming.

In [8], a 5-year survival model is proposed based on a Bayesian network for cancer patients with second primary cancer. By combining eleven cancer databases, it gathered 7845 patients' data. It used synthetic minority over-sampling techniques (SMOTE) to balance the dataset to combat unbalancing of the classes. Then artificial neural networks, support vector machines (SVM), logistic regression (LR), and the proposed method are used to report results. The results show that SMOTE boosts the sensitivity, and the proposed method outperforms other classifiers.

Trees have shown that they can be robust in many situations [9]. And since clinical data are related to biological systems, they are complex [10], they should be treated with robust classifiers to ensure decent results, or the results should be compared with tree-based classifiers [11]. Tree-based classifiers are also more robust against imbalance in data [12], and in [13] ensemble tree-based classifiers show better results in dealing with imbalance classification.

In [14], different data mining methods are examined to assess the colorectal cancer prediction of the SEER dataset. All of the methods are compared with TNM staging. Feature selection of each classifier was made using either backward step-wise feature selection or a genetic algorithm. Their study showed that the decision tree algorithm was the most accurate in predicting the survival rate of colorectal cancer patients. Similarly, Another study [15], proposed a two-stage model based on tree ensembles to predict the survival of patients with advanced-stage colorectal cancer. At this stage survival data are asymmetric due to the survival probability of 10%; therefore, a model is designed to provide satisfactory results despite this asymmetry. The model of the classification system is obtained by randomly choosing from the larger class with a smaller class size and training the system with a large number of symmetric models, which is finally determined by voting the classification result. The model showed better accuracy than traditional statistical.

In [16], an ensemble data mining approach was applied for survival prediction in lung cancer patients from the National Cancer Institute's Surveillance, Epidemiology, and End Results (SEER) dataset. The ensemble voting classifier in this paper achieved the best performance for 5-year survival prediction. The study showed that the ensemble approach improved the accuracy of prediction compared to individual models. A similar approach was taken by [17], who used ensemble data mining along with SMOTE for balancing to develop a colon cancer survival prediction model.

On the other hand, in [18], by using a deep neural network and by finding an optimal number of layers from 1 to 8, it obtained acceptable results compared to the two widely used random forest (RF) and LR methods. This paper used the data of 188,000 colon cancer patients from the SEER dataset and the best AUC was achieved with five hidden layers at about 0.88, showing its potential as a reliable method for predicting survivability in colon cancer patients.

In [19], they used a dataset from five European clinical trials for rectal cancer to develop models and nomograms to predict local recurrence, distant metastasis, and 5-year survival. Models and nomograms were based on Cox regression. The c-index for local recurrence, distant metastasis, and 5-year survival was 0.68, 0.73, and 0.70, respectively. In [20], a Cox regression method was used for calculating rectal cancer conditional survival. A dataset consisting of 22,610 patients with rectal adenocarcinoma from the SEER database was used in this paper. It was shown that 5-year survival time in all stages except stage I significantly improved. In [21], a multivariate Cox proportional hazards survival model was applied for rectal cancer patients.

In [22], authors proposed a hybrid machine learning approach for predicting the survival of patients with prostate cancer data from SEER database. Their model combined Cox regression and decision tree algorithms to improve prediction accuracy, and a cluster centroid under-sampling approach was applied to balance the data. The results showed that XGboost outperformed other classifiers in binary classification, while SVM achieved the best performance in the three-class type.

Cancer survival prediction is a crucial task, but many models suffer from low sensitivity due to the use of simple classification methods on imbalanced data. As pointed out in [23], the characteristics of the data play a vital role in improving prediction performance. To address the issue of imbalanced classification problems, many papers have proposed using oversampling or under-sampling techniques. Although these methods can improve classification performance to some extent, simple oversampling or the use of classifiers without addressing the imbalance of the data may lead to useless results.

In our study, we pre-processed data based on statistical tests, analyzed correlogram for feature selection, and developed several oversampling, under-sampling, and combined techniques for highly imbalanced dataset classification which did not consider in state-of-the-art methods. To tackle classification problems for the minority class, we introduced a novel hybrid sampling method and provide a fair comparison between various tree-based classifiers on a unified framework for survival prediction on 1-, 3-, and 5-year SEER datasets. The data is highly imbalanced in the 1-year task, imbalanced in the 3-year task, and balanced in the 5-year task, so balancing was not applicable for 5-year survival. To overcome the limitations of oversampling and under-sampling techniques, we propose a model that combines their features while making their shortcomings less affecting to the classifiers' result.

## 2. Experimental Setup

A block diagram of our proposed method is shown in Fig. 1. Colorectal cancer data were pre-processed, then different balancing methods were used to overcome imbalanced data problems. Finally, different tree-based machine-learning methods were applied for classification.

### 1.2. Dataset and Pre-processing

We used the SEER database [24] gathered since 1973. Colon and rectum cancer records from 2010 to 2015 were imported. The age of patients at the time of diagnosis was asserted, and those who were not in the range of 18 to 85 were removed from the dataset. With these enhancements total of 103,885 records, we removed those who died of anything but colorectal cancer. Moreover, the study is focused on the adenocarcinoma form of cancer, so from histologic type ICD-O-3, only the adenocarcinoma type was selected, and others were removed.

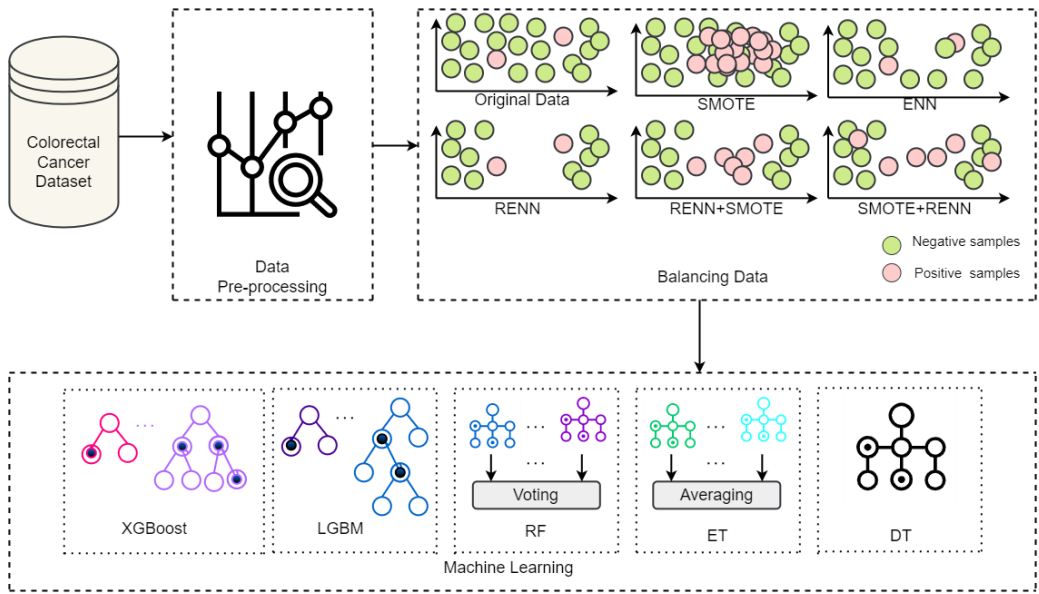

Figure 1. Block diagram representation of the proposed framework for colorectal cancer survival prediction.

Furthermore, records with missing values, coded or labeled Unknown, were also dropped. As a result of these modifications, the dataset size shrank to 42,764 records, and unbalancing in the data increased from 1:4 to 1:9 due to a high rate of missing values in patients who died under one year. This also indicates that missing values could be informative if filled with a tag, and the model would observe that pattern. This article chose the dropping policy instead of imputing missing values. The remaining records consist of about 8% with Grade I, 74% with Grade II, 15% with Grade III, and only 2% with Grade IV. Histologic Type consists of 78% *Adenocarcinoma, NOS*, 10% Adenocarcinoma in tubulovillous adenoma and others were put into the Others category. Sigmoid Colon occupies the most share by 21.5%, and the second and third most common places for cancer tumors in the dataset are Cesum and Rectum, NOS, respectively.

Liver metastases are mainly caused by colorectal cancer [25], so liver metastases information is vital to the outcome of such research. Nearly 10% have metastases to the liver, and others have no metastases at the liver.

Selected variables can be seen in Table 1. All variables were either categorical or were transformed into categories; for example, regional nodes examined is the total number of regional lymph nodes that were removed and examined; however, it was changed into five classes – 0, 10>,10-20, 20-30, and >30 Node Examined. Furthermore, some features were maintained to reduce their bins or simplify, like Median Household Income, whose categories were merged to half of its primary number of classes.

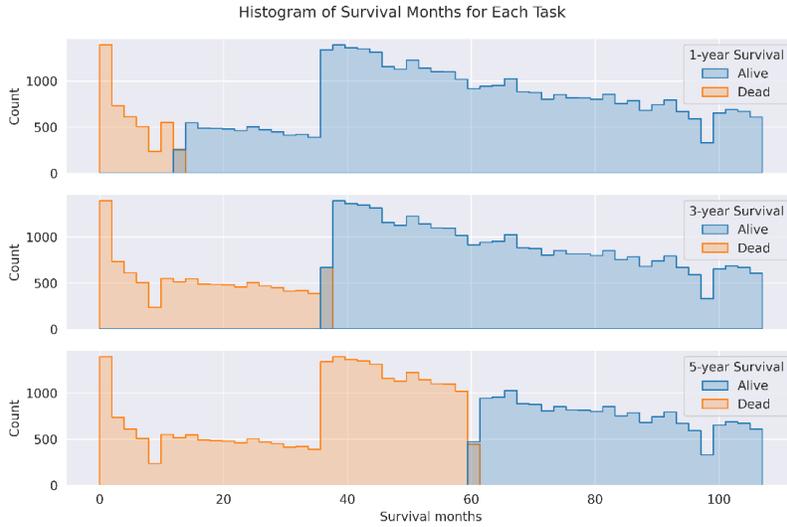

Figure 2. Histogram of survival months of patients

Table 1. Selected features from the colorectal cancer dataset

| Feature Name | Number of Categories | Name of Categories |
| --- | --- | --- |
| Sex | 2 | Male/Female |
| Grade | 4 | I/ II/ III/ IV |
| Race | 4 | W/B/AI/API |
| Primary Site | 8 | Sigmoid Colon/ Rectum/ Cecum/ Ascending Colon/ Rectosigmoid junction/ Transverse Colon/ Descending Colon/ Others |
| Histologic type ICDO-3 | 4 | >8389 & <8140 |
| Surgery of Primary Site | 3 | No Surgery, Tumor Destruction, Tumor Resection |
| CS Tumor Size | 5 | No Tumor/ 1-20/ 20-40/ 40-60/ 60-80 |
| Regional nodes examined | 5 | 0/ 1-9/ 10-30/ Others |
| Regional nodes positive | 3 | 0/ Any positive node/ Not Applicable |
| Median Household Income | 5 | <40k/ 40k-50/50-60/60-70/ >70k |
| Perineural Invasion Recode | 2 | Not Present/ Present |
| AJCC T, 6th ed | 5 | Tis/ T1/T2/T3/T4 |
| AJCC N, 6th ed | 3 | N0/N1/N2 |
| AJCC M, 6th ed | 2 | M0/M1 |
| CEA Pretreatment Interpretation Recode | 3 | Not Documented/ Positive/ Negative |
| CS extension | 4 | Categories are based on codes in the documentation of SEER |
| SEER Combined Mets at DX-liver | 2 | Positive/ Negative |
| CS lymph nodes | 6 | Categories are based on codes in the documentation of SEER |
| Number of in situ/malignant tumors | 3 | Categories are based on codes in the documentation of SEER |

We set our overall survival to 1-,3-, and 5-year, and since all data has enough follow-ups for this task, we map survival months to Survived and Not-Survived based on the task. A histogram of survival months of patients is shown in Fig. 2. As we can see, the data is highly imbalanced at 1-year survival, so about 90% of the records are labeled as Survived and 10% labeled as Not Survived.

After categorizing and reducing categories of all features, we performed an ANOVA test to remove non-significant elements from the dataset; in the result, Race, Sex, and Number of in situ/malignant tumors had a p-value higher than 0.05, while others had a p-value of less than 0.0001. As a result, non-significant features were removed.

We then performed a Crammer's V between the 16 remaining features to ensure that highly correlated features were not among the selected ones. The correlogram of the components is shown in Fig. 3.

In Fig. 3, most features correlate with each other except the component related to the income of the patient's family. This is due to the basis of the feature, which is socioeconomic, unlike other clinical features. Also, we can see that positive nodes and examined nodes have a high correlation because of their nature.

Correlations of features like the one between Collaborative Stage (CS), lymph nodes and AJCC 7 N should be considered as well. The N category is assigned a value of CS Lymph Nodes and the value of CS Site-Specific Factor 3, the Number of Positive Ipsilateral Axillary Lymph Nodes, so we selected AJCC 7 N and removed the other component.

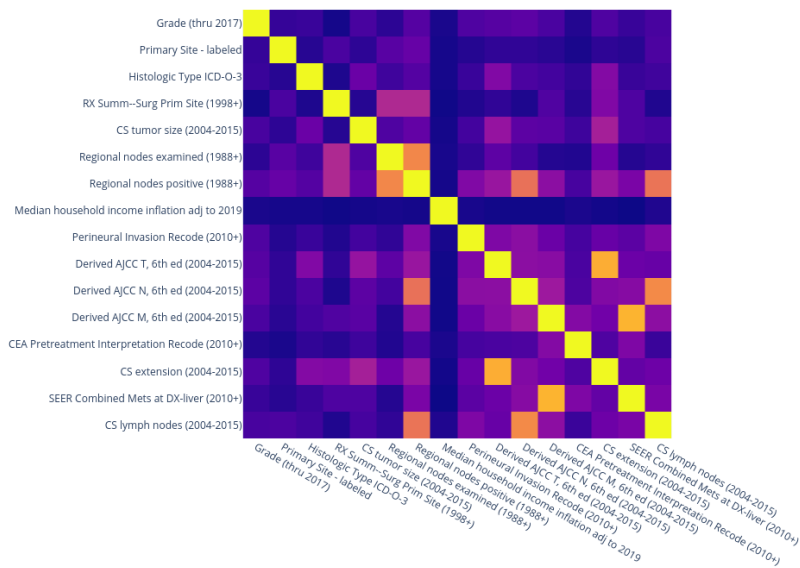

Figure 3. Correlogram of the features

## 2.2 Balancing Method

Imbalance data has a detrimental impact on machine learning algorithms. It biases classifiers into opting for the majority class as the classification's outcome, reducing precision or recall by misclassifying the minority class samples. Several methods are available to overcome this problem, including under-sampling and over-sampling. This study compared over-sampling and under-sampling methods, including Edited Nearest Neighbor (ENN) [26], Repeated Edited Nearest Neighbor (RENN) [27], and SMOTE [28], for data balancing. Also, we developed hybrid sampling approaches from SMOTE and RENN for balancing the data.

ENN under-samples the majority class by removing the samples with a different label than their k-nearest neighbors in the training set [26]. RENN repeats ENN algorithms until it draws all models with other brands in their k-nearest neighbors [27]. SMOTE oversamples the minority class by generating synthetic samples using a random sample of the minority class and its k-nearest neighbors [28]; however, since we have categorical features in the dataset, we have used a modified version that its output fits the categories of each class.

In highly imbalanced datasets, the challenge is simultaneously having good sensitivity and specificity, which can be reached by balancing the dataset or putting weights in the model to compensate for the gap in the class population. Here, we need better sensitivity in the models with no sampling method. We have tried to balance the dataset so that sensitivity is increased without a considerable loss in specificity.

In Fig. 4, 1000 samples were generated so that about 10% went to class A (purple) and 90% went to class B (yellow). And each of the samplers that we use in this paper were applied to those samples with 75% separability. Samplers' results are shown in Fig. 4. The number of samples varies based on the method. ENN drops 165 samples that are considered noisy by a five-neighbor metric. RENN removes more samples in comparison to ENN, which is plane. RENN reduces the dataset size to 728 from 1000, with about 100 more samples reduced than ENN. Unlike the under-sampling methods, SMOTE adds to the dataset size and increases the dataset's length to 1792 samples. The combined approach follows the RENN method by a SMOTE that reduces dataset size at first and then increases its size to 1248 samples. At last, SMOTE plus RENN is shown which result in the same number of samples as the input.

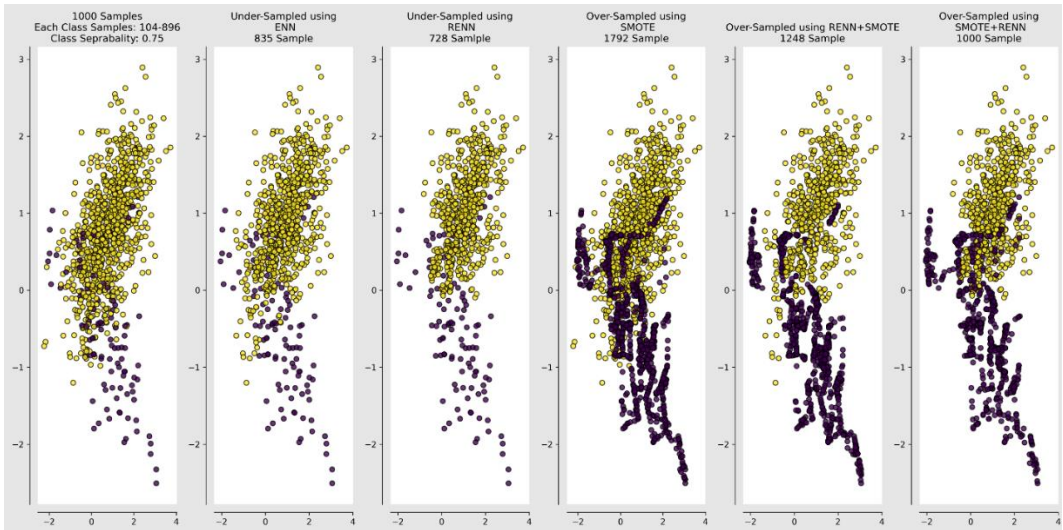

Figure 4. Comparison of sampling methods

Looking at the details of samples in each sampler, we can see that ENN and RENN remove a significant amount of data from the overlap area of classes and biases the outcome of the classification. Also, SMOTE strengthens patterns that are weak by making and generating numerous samples from them and as a result models predict falsely based on these patterns. RENN+SMOTE removes noisy data and then creates synthesis data. SMOTE+RENN draws data patterns as RENN does and generates a bundle of data from minority class in the overlap.

## 3.2. Classification

Among different simple machine learning algorithms, the decision tree (DT) [29] is more interpretable for clinical data processing since it can recognize non-linear patterns more effectively than other base learners. A Decision Tree is a supervised learning technique that can be used for classification and regression problems, but it is mainly preferred for solving classification problems. It is a tree-based classifier where internal nodes represent the features of a dataset, branches represent the decision rules, and each leaf node represents the outcome. Decision Trees usually mimic human thinking ability while making a decision, so it is easy to understand and can be beneficial for solving decision-related problems. However, it may have an overfitting issue, which can be resolved using ensemble methods.

Ensemble methods try to prevent overfitting and reduce the variance of prediction of its bias based on their structure. Extra trees (ET) and RF [30] are ensemble methods based on voting or averaging. RF creates numerous trees and trains each tree by a proportion on the train set. Then each tree estimates the input, and the RF prediction aggregates those predictions by averaging or voting. ET has the same structure, though it splits nodes based on random thresholds, resulting in even lower variance than RF— main extended DT versions, generally ensemble DT.

Boosting is an ensemble method where unused models are included to adjust the errors made by existing models. Models are included successively until no further advancements can be made. XGboost [31] is a more advanced method that opts for a reduction in bias via reducing the error by building new trees and aggregating the results. Gradient boosting is an approach in which unused models foresee the residuals or blunders of earlier models and are added together to create the ultimate prediction. It is called gradient boosting since it employs a gradient descent algorithm to decrease the loss when including new models. LGBM [32] is considered a rapid algorithm and the foremost utilized algorithm in machine learning for getting fast and high-precision results. LGBM develops trees vertically, whereas other algorithms develop trees horizontally, meaning Light GBM grows trees leaf-wise, while different algorithms develop level-wise. It'll select the leaf with max delta loss to conceive. When raising the same leaf, a Leaf-wise algorithm can diminish more upsets than a level-wise algorithm.

## 3. Results and Discussion

For more reliable and accurate analysis, we developed a 5-fold cross-validation in which 80% of data was used for training, and 20% was used for testing the performance of algorithms in each fold.

Several metrics, including accuracy, sensitivity, specificity, and F1-score, were computed as described in Equations (1-4) to evaluate the proposed method.

| | |
|---|---|
| $Accuracy = \dfrac{Tn + Tp}{Tn + Fp + Tp + Fn}$ | (1) |
| $Sensitivity = \dfrac{Tp}{Tp + FN}$ | (2) |
| $Specificity = \dfrac{Tn}{Tn + Fp}$ | (3) |
| $F1_{score} = \dfrac{1}{1 + \dfrac{Fn}{2Tp} + \dfrac{Fp}{2Tp}}$ | (4) |

In Equations (1-4), TP, TN, FP, and FN are represented as a true positive, true negative, false positive, and false negative, respectively. The performance of the proposed method is summarized in Table 2 and Table 3.

Table 2 reports the data of models with no sampling method for all three tasks. 1-year survival prediction task is highly imbalanced; in this regard, the accuracy and specificity are about 90%, while the sensitivity and f-score are near 10%, which means that the false negative rates are too high for an acceptable classification. The best sensitivity and f-score for the 1-year survival prediction task were obtained using the XGboost algorithm, which shows the robustness of boosting algorithm using trees to counter imbalance data compared to other classifiers used in this article. The 3-year survival prediction data in Table 2 is still unbalanced, although milder than the 1-year survival prediction. The accuracy and specificity in this task are degraded by about 10% and 5%, respectively, while the sensitivity and f-score are increased by about 30%. The best sensitivity and f-score in this task were achieved by LGBM, which suggests that LGBM is a better classifier for mild imbalanced data classification among the used classifiers. The 5-year survival prediction was balanced, and the best accuracy, sensitivity, and f-score were achieved by LGBM, which illustrates this classifier's strength in classifying clinical outcomes.

Table 2. Performance comparison between different classifiers in the proposed 1-year, 3-year, and 5-year survival prediction.

| Model | Survival time | Accuracy | Sensitivity | Specificity | F-score |
|---|---|---|---|---|---|
| DT | 1-year | 90.63% | 6.74% | 99.36% | 11.96% |
|  | 3-year | 81.91% | 40.28% | 94.14% | 50.24% |
|  | 5-year | 60.55% | 58.10% | 63.76% | 62.57% |
| EDT | 1-year | 90.61% | 2.80% | 99.57% | 5.33% |
|  | 3-year | 82.10% | 38.06% | 94.94% | 48.98% |
|  | 5-year | 60.94% | 60.96% | 60.92% | 63.91% |
| RF | 1-year | **90.70%** | 7.30% | 99.38% | 12.88% |
|  | 3-year | 82.16% | 38.90% | 94.77% | 49.60% |
|  | 5-year | 61.02% | 62.09% | 59.62% | 64.38% |
| XGboost | 1-year | 90.22% | 10.67% | 98.50% | 17.06% |
|  | 3-year | 81.57% | 38.74% | 94.05% | 48.68% |
|  | 5-year | 60.22% | 62.49% | 57.23% | 64.06% |
| LGBM | 1-year | 90.62% | 8.04% | 99.21% | 13.91% |
|  | 3-year | 82.22% | 40.66% | 94.34% | 50.80% |
|  | 5-year | 61.16% | 63.03% | 58.70% | 64.81% |

In Fig. 4, sensitivity and F-score of all methods are demonstrated. As we can see in Fig. 4, the results of 1-year survival are poor and by having more balance in the data, they get better. Also, variations in 1-year task's results are higher between classifiers in compare to other tasks.

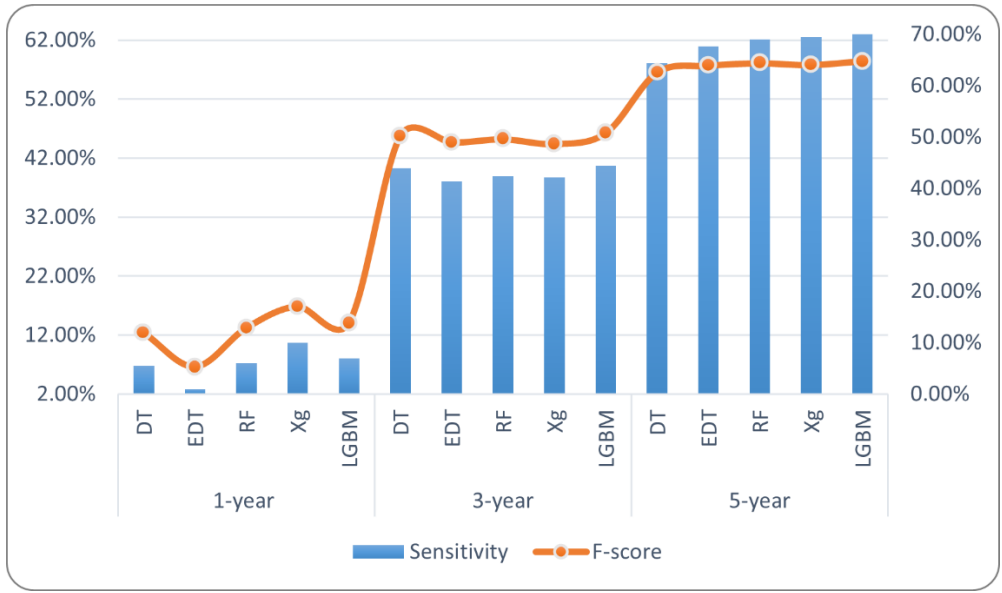

Figure 4. Sensitivities (left axis) and F-Score (right axis) of all tasks for the classifiers with no sampling method

In Table 3, the results of the 1-year survival classification are reported in terms of accuracy, sensitivity, specificity, and F1-score. This paper compares SMOTE, ENN, RENN, and combinations of RENN and SMOTE as sampling methods using different tree-based algorithms. In 1-year survival, the best sensitivity is achieved by a RENN and SMOTE orderly pipeline with LGBM, resulting in 15% higher sensitivity than RENN with LGBM. The sampling method results in the best sensitivity in other classifiers as well. This indicates that the process is robust for increasing the true positive rate in the model. Also, the models' sensitivity increased significantly compared to no-sampling models of Table 2, which are less than 11% for 1-year survival—in 1-year survival combining SMOTE and RENN with LGBM, which had a sensitivity of about 0.08 in the non-sampling method, improved its capability of predicting minority class 9 times better, reaching 0.72. This significant improvement is also nearly 15% higher sensitivity than SMOTE or RENN sampling method alone. This indicates the model's capability to deal with rare event predict imbalanced datasets. Also, the procedure is robust on all tree-based classifiers, showing a range of 0.12-0.27 increase in F1-score using different methods. Sensitivities and F-score of three top sampling methods including the novel method comparing with no sampling method are shown in Fig. 5, and as we can see the significant improvement in the sampling methods is quite notable. Moreover, the significant rise in the sensitivity does not decrease F-score. The highest sensitivity belongs to RENN+SMOTE across classifiers which has a slight lower F-score in compare to RENN. This less improvement in F-Score is a good tradeoff for about 0.2 rise in sensitivity.

Table 3. Performance comparison between different sampling models and classifiers in the proposed 1-year survival prediction.

| Model | Sampling | Accuracy | Sensitivity | Specificity | F-score |
|---|---|---|---|---|---|
| DT | SMOTE | 71.39% | 59.23% | 72.66% | 28.07% |
|  | ENN | 89.40% | 24.59% | 96.15% | 30.44% |
|  | RENN | 82.72% | 48.81% | 86.25% | 34.75% |
|  | RENN+SMOTE | 65.78% | 70.99% | 65.23% | 28.11% |
|  | SMOTE+RENN | 80.97% | 45.91% | 84.63% | 31.27% |
| EDT | SMOTE | 78.29% | 54.00% | 80.82% | 31.91% |
|  | ENN | 89.88% | 23.15% | 96.82% | 30.12% |
|  | RENN | 84.31% | 47.00% | 88.19% | 36.09% |
|  | RENN+SMOTE | 68.87% | 70.49% | 68.70% | 29.91% |
|  | SMOTE+RENN | 83.35% | 42.61% | 87.59% | 32.54% |
| RF | SMOTE | 78.33% | 49.26% | 81.36% | 30.00% |
|  | ENN | 89.40% | 26.58% | 95.93% | 32.09% |
|  | RENN | 82.66% | 51.56% | 85.90% | **35.92%** |
|  | RENN+SMOTE | 67.57% | 70.89% | 67.23% | 29.18% |
|  | SMOTE+RENN | 82.97% | 39.75% | 87.47% | 30.56% |
| XGboost | SMOTE | 76.13% | 49.90% | 78.86% | 28.26% |
|  | ENN | 88.59% | 28.09% | 94.89% | 31.70% |
|  | RENN | 81.48% | 51.46% | 84.60% | 34.37% |
|  | RENN+SMOTE | 60.91% | 71.25% | 59.83% | 25.57% |
|  | SMOTE+RENN | 79.68% | 44.32% | 83.37% | 29.14% |
| LGBM | SMOTE | 76.24% | 50.59% | 78.90% | 28.64% |
|  | ENN | 88.91% | 29.70% | 95.07% | 33.55% |
|  | RENN | 78.73% | 57.59% | 80.93% | 33.79% |
|  | RENN+SMOTE | 63.06% | **72.30%** | 62.10% | 26.95% |
|  | SMOTE+RENN | 82.34% | 41.27% | 86.62% | 30.58% |

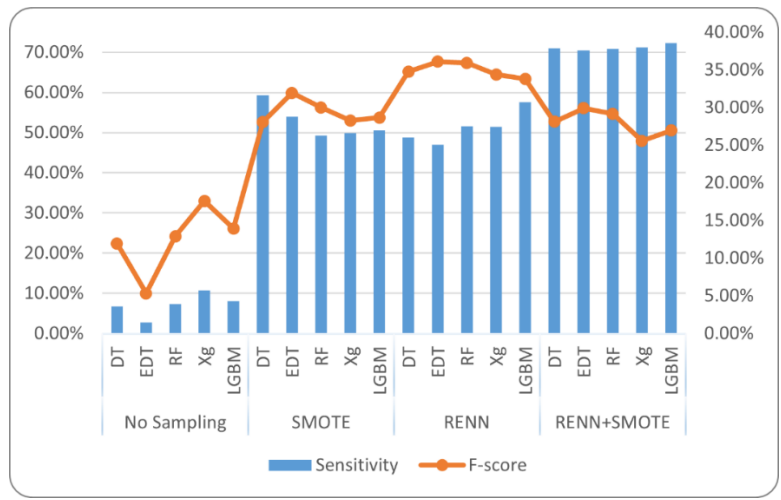

Figure 5. Sensitivity and F-score of 1-year survival for each classifier using No-sampling and 3 top samplers

Table 4. Performance comparison between different sampling models and classifiers in the proposed 3-year survival prediction.

| Model | Sampling | Accuracy | Sensitivity | Specificity | F-score |
|---|---|---|---|---|---|
| DT | SMOTE | 74.38% | 65.05% | 77.10% | 53.41% |
| | ENN | 79.86% | 54.37% | 87.29% | 54.92% |
| | RENN | 68.87% | 74.71% | 67.20% | 52.00% |
| | RENN+SMOTE | 68.89% | 74.67% | 67.21% | 52.00% |
| | SMOTE+RENN | 80.27% | 48.14% | 89.64% | 52.41% |
| EDT | SMOTE | 76.83% | 64.12% | 80.53% | 55.54% |
| | ENN | 80.69% | 55.10% | 88.15% | 56.29% |
| | RENN | 68.14% | 77.42% | 65.43% | 52.31% |
| | RENN+SMOTE | 68.88% | 76.77% | 66.58% | 52.69% |
| | SMOTE+RENN | 80.60% | 49.52% | 89.66% | 53.54% |
| RF | SMOTE | 76.74% | 63.08% | 80.72% | 55.04% |
| | ENN | 79.88% | 58.65% | 86.07% | **56.82%** |
| | RENN | 65.63% | 79.53% | 61.58% | 51.09% |
| | RENN+SMOTE | 66.00% | 79.01% | 62.21% | 51.20% |
| | SMOTE+RENN | 80.28% | 49.53% | 89.25% | 53.14% |
| XGboost | SMOTE | 75.48% | 61.42% | 79.58% | 53.07% |
| | ENN | 78.90% | 57.91% | 85.01% | 55.33% |
| | RENN | 64.79% | 78.21% | 60.87% | 50.06% |
| | RENN+SMOTE | 64.99% | 78.29% | 61.11% | 50.24% |
| | SMOTE+RENN | 78.55% | 49.36% | 87.06% | 50.95% |
| LGBM | SMOTE | 76.62% | 63.00% | 80.59% | 54.89% |
| | ENN | 79.13% | 61.27% | 84.35% | 57.00% |
| | RENN | 63.44% | **80.81%** | 58.38% | 49.95% |
| | RENN+SMOTE | 63.83% | 80.65% | 58.93% | 50.16% |
| | SMOTE+RENN | 79.78% | 50.61% | 88.29% | 53.06% |

Table 4 reports models with sampling methods for 3-year survival. In 3-year survival, the best sensitivity is achieved using RENN as the sampler and LGBM as the classifier. The increase in sensitivity is about 40%, double the true positive rate with no sampler, showing that minority class would be predicted more accurately by sampling. Also, the f-score is not significantly dropped; even we can see an increase in the results of the models that RF using ENN as a sampler has a considerably higher f-score than models with no sampling. As we can see in Fig. 6, the performance of RENN and the hybrid sampler does not differ, this is due to the fact that 3-year survival is much less imbalance in compare to 1-year task. The structure of the proposed method focuses on highly imbalanced datasets, as a result the metrics for it in 3-year survival are not better than RENN.

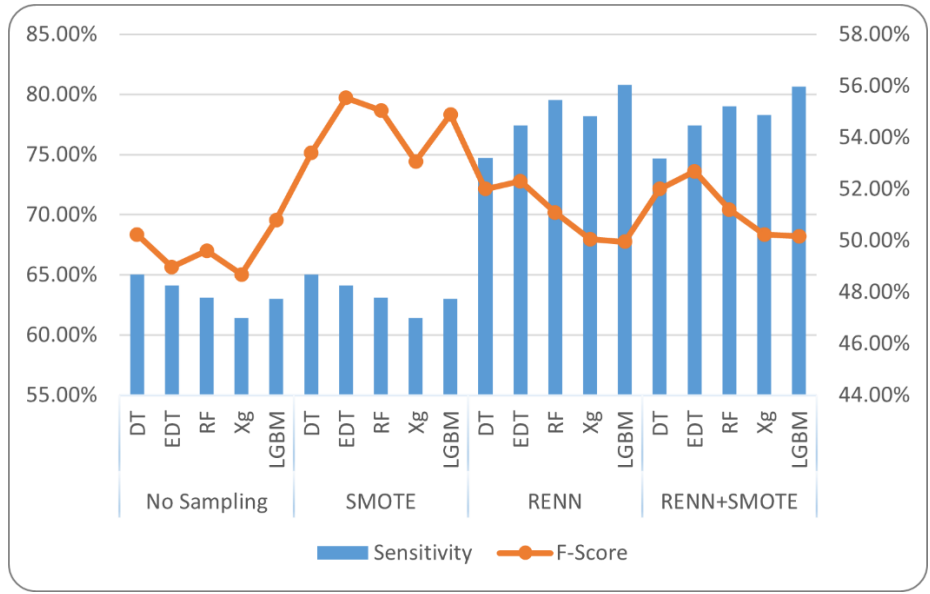

Figure. 6. Sensitivity and F-score of 3-year survival for each classifier using No-sampling and 3 top samplers

## 4. Conclusions

The field of survival prediction for colorectal cancer has seen significant advancements, with researchers proposing methods and critical features for assessing the survivability of this disease. Clinical data along with machine learning plays a pivotal role in analyzing colorectal cancer, providing invaluable insights into patients' conditions. However, the imbalance of disease outcomes poses a challenge for machine learning models, as they often struggle to handle such imbalances. In our study, we built models for predicting 1-, 3-, and 5-year survival of colorectal cancer from SEER data, and compared the sampling models and tree-based classifiers with each other.

We investigated the use of oversampling techniques for robust predictions; however, in the case of minority oversampling alone may not sufficiently improve prediction outcomes due to the high volume of synthetic data created. Conversely, under-sampling may result in the loss of valuable training data while attempting to balance the classes. To overcome these limitations, we employed Synthetic Minority Over-sampling Technique (SMOTE) to generate synthetic data and balance the classes, then the Repeated Edited Nearest Neighbor (RENN) algorithm to remove noisy data. The proposed novel approach yielded a notable enhancement in sensitivity while demonstrating minimal impact on other metrics such as F-score, thus establishing its suitability as a sampling technique. Our study underscores the significance of addressing data imbalance in survival prediction for colorectal cancer. Through the integration of SMOTE and RENN, we achieved improved sensitivity in predicting 1-year survival. However, it should be noted that due to its higher computational cost, this method is not recommended for mildly imbalanced datasets, as observed in our 3-year survival prediction scenario. This indicates that using the proposed method in highly imbalanced dataset can be invaluable while it could result in more computational cost in mild imbalance datasets.

Future studies can explore additional techniques such as handling missing values, feature selection, dimension reduction, and utilizing alternative machine learning methods to further enhance the predictive performance. Specifically, comparing different missing values handling methods, including imputing them with median, iterative imputer, and nearest neighbor imputer, would be valuable for future investigations.

## Conflict of Interest

Authors Sadegh Soleimani, Mahsa Bahrami, and Mansour Vali declare no conflict of interest.

## Human Studies/Informed Consent

The authors carried out no human studies for this article.

## Animal Studies

The authors carried out no animal studies for this article.